\documentclass[a4paper,twoside]{article}

\usepackage{epsfig}
\usepackage{subcaption}
\usepackage{calc}
\usepackage{amssymb}
\usepackage{amstext}
\usepackage{amsmath}
\usepackage{amsthm}
\usepackage{multicol}
\usepackage{pslatex}
\usepackage{apalike}
\usepackage[bottom]{footmisc}
\usepackage[dvipsnames]{xcolor}
\usepackage{multirow}
\usepackage{booktabs}
\usepackage[labelformat=simple]{subcaption}

\usepackage{SCITEPRESS}     

\begin{document}
\def\eg{\emph{e.g. }}
\def\Eg{\emph{E.g. }}
\def\etal{\emph{et al. }}
\def\ie{\emph{i.e. }}

\title{SynMotor: A Benchmark Suite for Object Attribute Regression and Multi-task Learning}

\author{\authorname{Chengzhi Wu\sup{1}, Linxi Qiu\sup{1}, Kanran Zhou\sup{1}, Julius Pfrommer\sup{2,3} and Jürgen Beyerer\sup{3}}
\affiliation{\sup{1}Institute for Anthropomatics and Robotics, Karlsruhe Institute of Technology, Karlsruhe, Germany}
\affiliation{\sup{2}Fraunhofer Center for Machine Learning, Karlsruhe, Germany}
\affiliation{\sup{3}Fraunhofer Institute of Optronics, System Technologies and Image Exploitation IOSB, Karlsruhe, Germany}
\email{chengzhi.wu@kit.edu, \{linxi.qiu, kanran.zhou\}@student.kit.edu, \{julius.pfrommer, juergen.beyerer\}@iosb.fraunhofer.de}
}



\abstract{In this paper, we develop a novel benchmark suite including both a 2D synthetic image dataset and a 3D synthetic point cloud dataset. Our work is a sub-task in the framework of a remanufacturing project, in which small electric motors are used as fundamental objects. Apart from the given detection, classification, and segmentation annotations, the key objects also have multiple learnable attributes with ground truth provided. This benchmark can be used for computer vision tasks including 2D/3D detection, classification, segmentation, and multi-attribute learning. It is worth mentioning that most attributes of the motors are quantified as continuously variable rather than binary, which makes our benchmark well-suited for the less explored regression tasks. In addition, appropriate evaluation metrics are adopted or developed for each task and promising baseline results are provided. We hope this benchmark can stimulate more research efforts on the sub-domain of object attribute learning and multi-task learning in the future.}

\onecolumn \maketitle \normalsize \setcounter{footnote}{0} \vfill

\section{\uppercase{Introduction}}
\label{sec:introduction}
Machine learning researchers have developed tremendous inventive network models and algorithms during the past decade. In parallel, a relatively small number of benchmarks have been developed for evaluating and comparing the performance of various models. Datasets and benchmarks play important roles in the development of neural networks and drive research in more challenging directions. A good dataset can boost the development of a certain computer vision domain, \eg, ImageNet \cite{Deng2009ImageNetAL} to image classification, PASCAL VOC \cite{Everingham2009ThePV} and COCO \cite{Lin2014MicrosoftCC} to image detection and segmentation, KITTI \cite{Geiger2012AreWR} to point cloud detection, or ShapeNet \cite{Chang2015ShapeNetAI} and S3IDS \cite{Armeni20163DSP} to point cloud segmentation. 
However, among all the computer vision tasks, the task of attribute regression is less explored due to the scarcity of suitable benchmarks. Current attribute learning methods mostly focus on outdoor pedestrians \cite{Li2015MultiattributeLF}\cite{Li2018PoseGD} or human facials \cite{Sarafianos2018DeepIA}\cite{Kalayeh2017ImprovingFA}. The attributes in those datasets are mostly binary (\eg gender, with glasses or not), and only a few attributes are continuous variables (\eg age). On the other hand, regression models with neural networks are mostly used for non-vision data. We think it would be interesting if we could contribute to bridging the gap between attribute learning and regression for the computer vision community.

\begin{figure*}[t]
    \centering
    \includegraphics[width=1.0\linewidth]{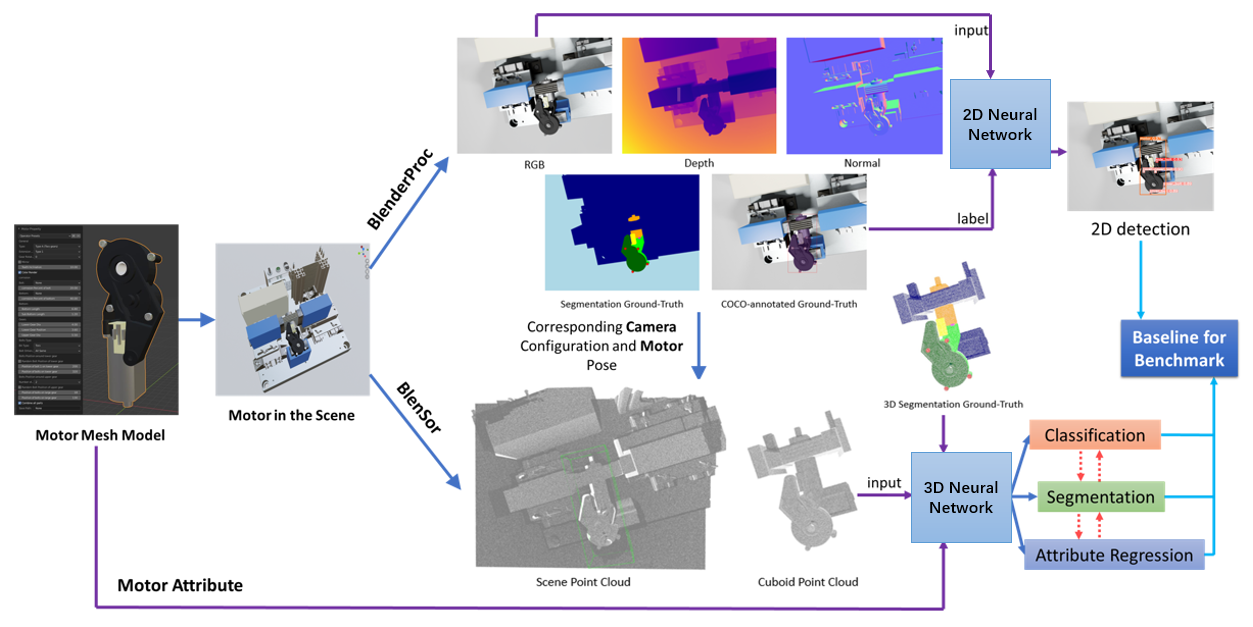}
    \caption{The framework of our work. Synthetic datasets are generated and task metrics are defined. Baseline results of 2D detection and 3D classification, segmentation, and attribute regression are provided.}
    \label{fig:framework}
\end{figure*}

In this paper, we propose SynMotor, a benchmark that gives the possibility to perform object attribute regression. Multi-task learning and multi-modal learning are also possible with the provided dataset. The benchmark originates from a sub-task within our manufacturing project which aims at the automatic disassembly of small electric motors. A mesh model dataset is first generated with a carefully developed Blender addon, in which the motor specifications are saved as object attributes. 
Subsequently, synthetic motor datasets of both 2D image dataset and corresponding 3D point cloud dataset are created for deep learning purposes. Apart from the object attributes ground truth, the ground truth label of common computer vision tasks including detection, classification, and segmentation are also provided. Those labels are generated automatically along with the image or point cloud generation, with no manual annotation required. On the other hand, the metrics for common computer vision tasks are already well-developed, while developing metrics for object attribute regression is more difficult since the metrics need to be attribute-oriented. In our benchmark, we have designed several metrics for attribute regression given the motor attributes. Baseline results are also provided in the following experiments with some widely recognized networks. An illustrative figure of our framework is given in Figure \ref{fig:framework}.

The remainder of this paper is structured as follows: Section \ref{sec:relatedWork} summarizes the state-of-the-art of 3D dataset creation, multi-task learning, and attribute learning. Section \ref{sec:dataset} shows a pipeline of creating our synthetic dataset. Section \ref{sec:tasksAndMetrics} describes the tasks and the developed metrics. Section \ref{sec:baseline} gives baseline results with some widely recognized network models. Finally, Section \ref{sec:conclusion} summarizes presented results and discusses future work.

\section{\uppercase{Related work}}
\label{sec:relatedWork}
\textbf{3D dataset.}
For 3D object dataset, ModelNet \cite{Wu20153DSA} builds a huge dataset of 3D CAD models and provides the ModelNet40 benchmark for tasks including 3D shape classification and retrieval. Another similar work of ShapeNet \cite{Chang2015ShapeNetAI} provides more detailed semantic annotations, its subsequent work of PartNet \cite{Mo2019PartNetAL} additionally offers fine-grained semantic segmentation information for a subset of the models. \cite{Trem18} creates a dataset for object pose estimation. A large dataset of 3D-printing models is provided in Thingi10K \cite{Zhou2016Thingi10KAD}, while a more recent ABC dataset \cite{Koch2019ABCAB} collects over 1 million CAD models including many mechanical components.
Regarding 3D scenes, KITTI \cite{Geiger2012AreWR} uses a vehicle-mounted platform outfitted with a variety of sensors to capture and record road information. Data for autonomous driving tasks like 3D object detection and 3D tracking are collected. While \cite{Ros16} and \cite{Kha19} generate synthetic datasets for the segmentation and detection of objects in virtual urban scenes, \cite{Hoa20} generates images from virtual garden scenes. SynthCity \cite{Gri19} generates point clouds of urban scenes using Blender. \cite{Pie19} also uses Blender but for the generation of point clouds of historical objects. For indoor scenes, SUN-RGBD \cite{Song2015SUNRA}, S3DIS \cite{Armeni20163DSP}, and ScanNet \cite{Dai2017ScanNetR3} use different cameras to scan rooms to get 3D indoor point cloud dataset.

\begin{figure}[t]
    \centering
    \begin{subfigure}[b]{0.46\linewidth} 
        \centering
        \raisebox{0.3cm}{
        \includegraphics[width=\linewidth,trim=2 2 2 2,clip]{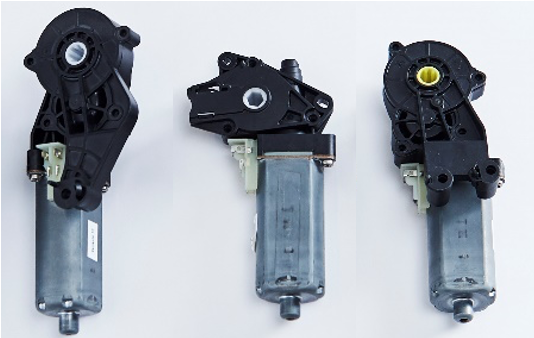}}
        \caption{\vspace{0.2cm}}
        \label{fig:realMotor}
    \end{subfigure}
    \hspace{0.2cm}
    \begin{subfigure}[b]{0.46\linewidth} 
        \centering
        \includegraphics[width=\linewidth,trim=2 2 2 2,clip]{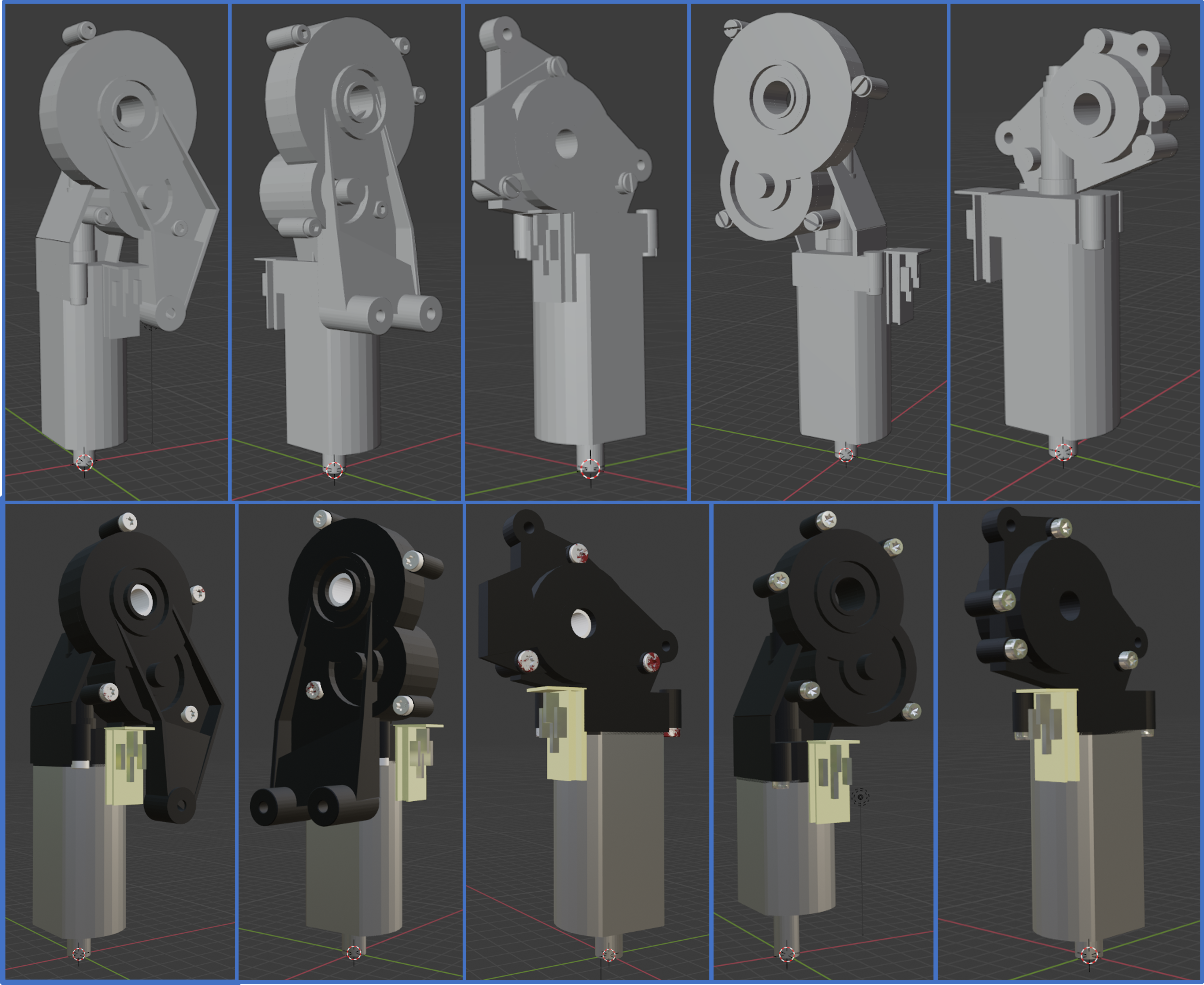}
        \caption{\vspace{0.2cm}}
        \label{fig:motorDemo1}
    \end{subfigure}
    \begin{subfigure}[b]{0.7\linewidth} 
        \centering
        \includegraphics[width=\linewidth,trim=2 2 2 2,clip]{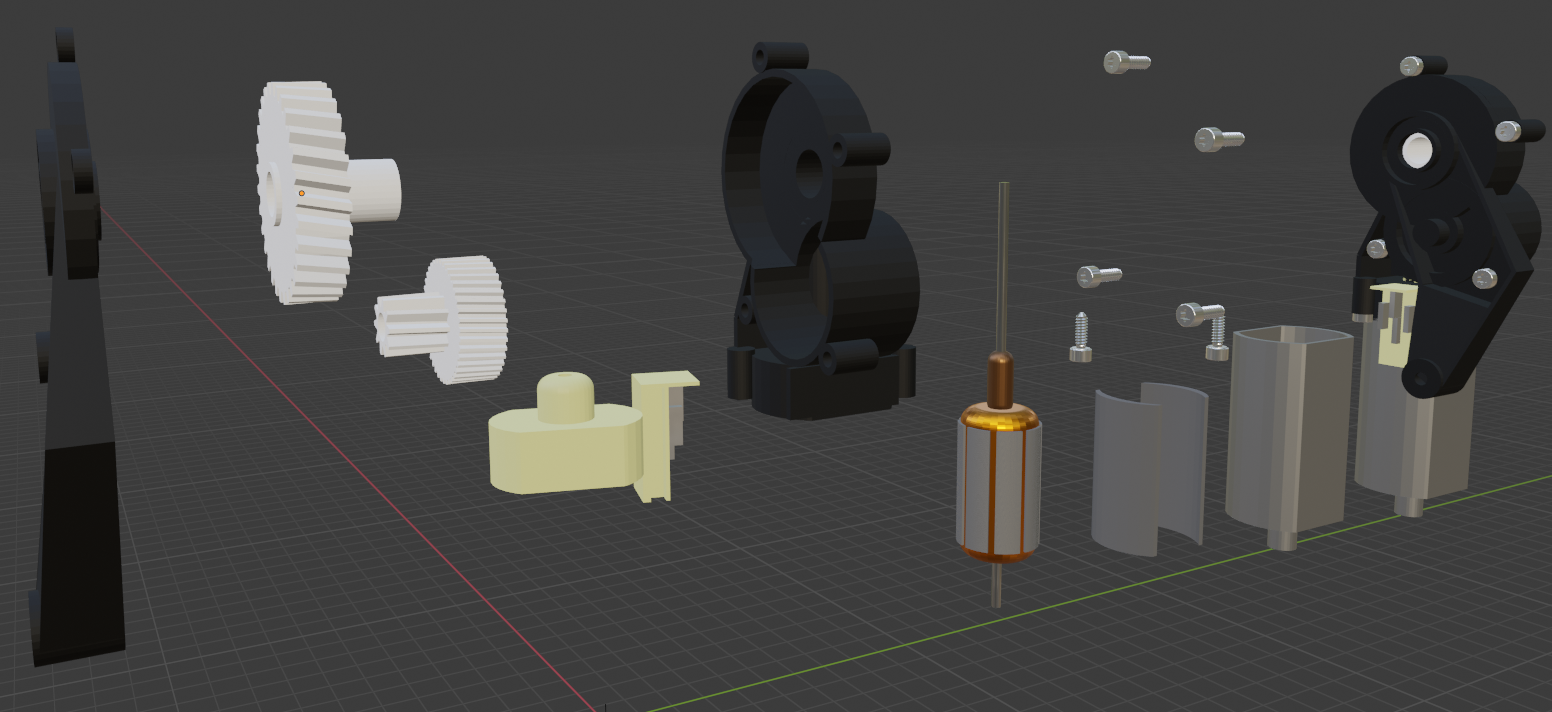}
        \caption{}
        \label{fig:motorDemo2}
    \end{subfigure}
    \caption{Real-world motors (a) and generated demo motors. (b) Upper row: no textures added; bottom row: textures added and rendered. (c) An explosion figure of a generated motor. The original assembled motor model is also shown at the rightmost.  \vspace{-0.cm}}
    \label{fig:motorDemo}
\end{figure}

\vspace{0.2cm}

\textbf{Multi-task learning.}
Overfeat \cite{Sermanet2014OverFeatIR} trains on classification, localization, and detection tasks simultaneously using a single shared convolutional network.
\cite{Eigen2015PredictingDS} implements depth prediction, normal estimation, and semantic labeling using a single multi-scale convolutional network.
\cite{Kendall2018MultitaskLU} optimizes the weight of loss during multi-task training according to the uncertainty of each task, thereby realizing the simultaneous learning of classification and regression tasks of different orders of magnitude.  MultiNet \cite{Teichmann2018MultiNetRJ} and UberNet \cite{Kokkinos2017UberNetTA} offer methods for jointly performing classification, detection, and semantic segmentation using a unified architecture and achieving very efficient results. The above approaches work on the fundamental principle of a global feature extractor comprised of convolutional layers shared by all tasks and a different output branch for each task. In contrast to previous approaches, the strategy used in  \cite{Misra2016CrossStitchNF} and \cite{Gao2019NDDRCNNLF} is to have a separate network for each task while exchanging parameters between their parallel layers. Ruder \cite{Ruder2019LatentMA} divides each layer into task blocks and information sharing blocks of previous layers, and the input of each layer is a linear combination of these two blocks, so that when learning, you can choose to focus more on the relevance of the task or the task itself. PAD-Net \cite{Xu2018PADNetMG} uses multi-task learning making preliminary predictions for depth, scene, and surface normal estimation, then combining these predictions to obtain the refined depth and scene parsing results.

\vspace{0.2cm}

\textbf{Attribute learning.}
Attribute learning refers to the process of discovering relevant attributes based on known logical principles. Visual attribute recognition has become an important research area due to its high-level semantic information. Previous work like \cite{Russakovsky2010AttributeLI} establishes transfer learning by learning visual attributes such as colors, shapes, or textures of images, and makes connections between semantically unrelated categories. Attribute learning has a wide range of applications in the recognition of pedestrians. \cite{Li2015MultiattributeLF} suggests two deep learning models for surveillance scenarios to recognize based on a single attribute and multiple attributes respectively. PGDM \cite{Li2018PoseGD} analyses the salient features of the pedestrian's body structure, it is helpful for the recognition of pedestrian attributes. DIAA \cite{Sarafianos2018DeepIA} utilizes multi-scale visual attention and weighted focal loss for person attribute recognition. With the widespread popularity of posting selfies in social software, the practice of recognizing facial attributes is growing, \cite{Liu2015DeepLF} provides a cascaded deep learning framework for jointly predicting attributes and localizing faces. \cite{Kalayeh2017ImprovingFA} leverages the information obtained from semantic segmentation to improve facial feature prediction, \cite{Liu2018ExploringDF} obtains equivalent discriminative ability for face attribute recognition on CelebA and LFWA \cite{Huang2008LabeledFI} datasets by learning disentangled but complementing face features with minimal supervision.

\begin{figure}[t]
    \centering
    \begin{subfigure}[b]{0.5\linewidth} 
        \centering
        \includegraphics[width=\linewidth,trim=2 2 2 2,clip]{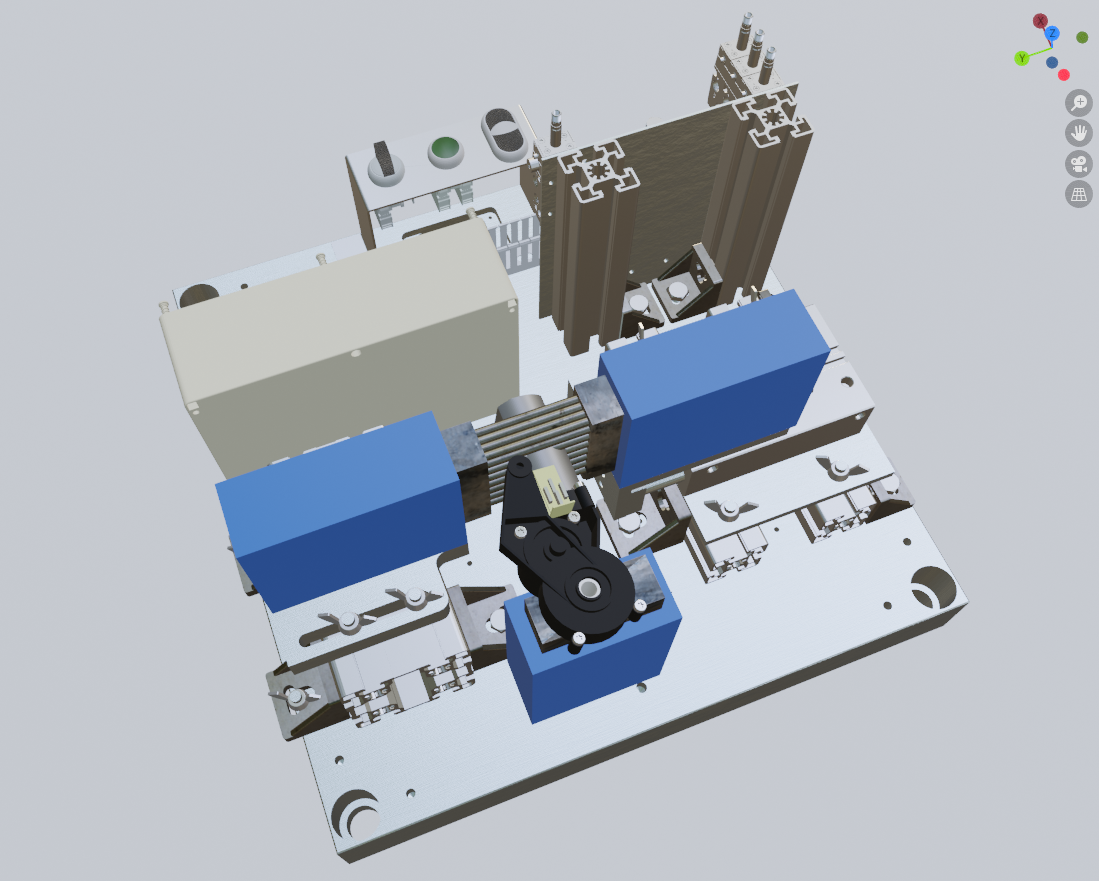}
        \caption{\vspace{0.2cm}}
        \label{fig:scene}
    \end{subfigure}
    \hspace{0.2cm}
    \begin{subfigure}[b]{0.45\linewidth} 
        \centering
        \includegraphics[width=\linewidth,trim=2 2 2 2,clip]{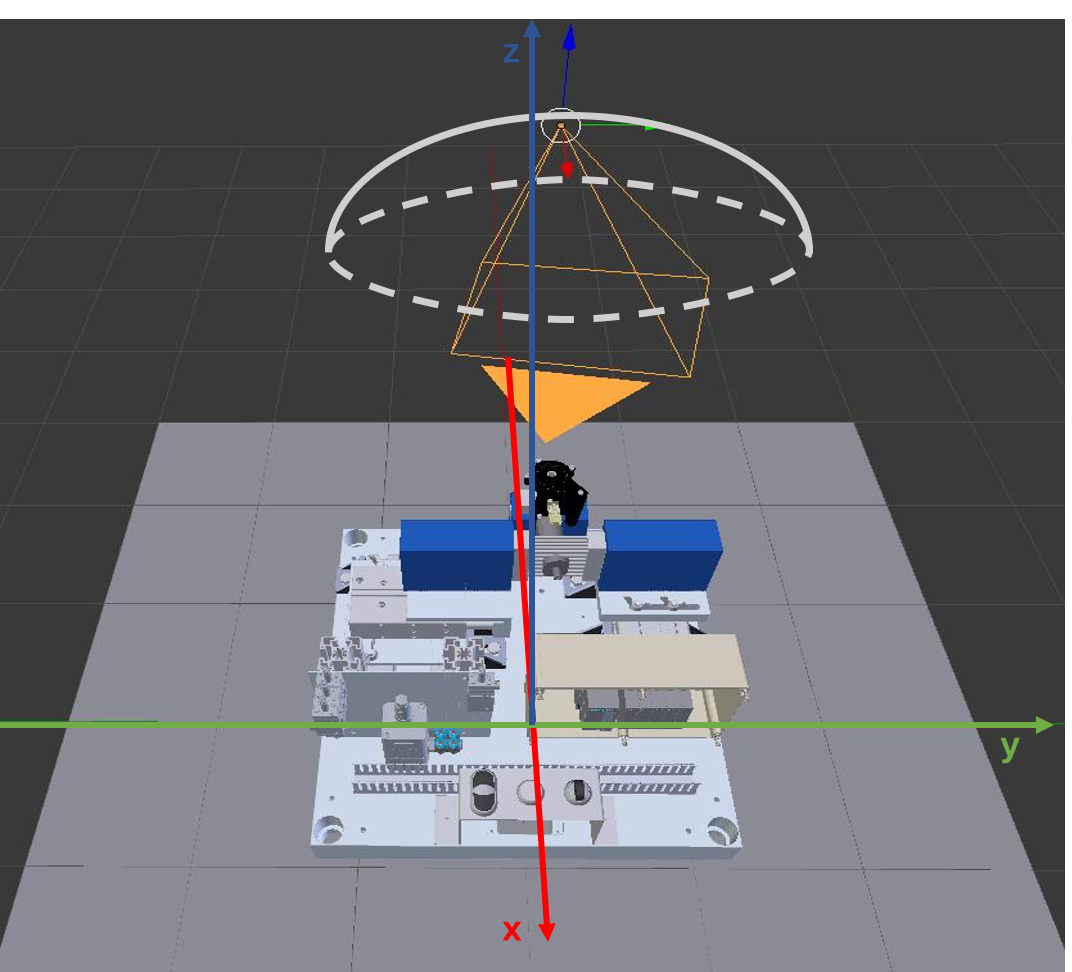}
        \caption{\vspace{0.2cm}}
        \label{fig:camera}
    \end{subfigure}
    \caption{(a) Simulated scene in Blender, with a clamping system. (b) Camera setting.} 
    \label{fig:sceneCamera}
\end{figure}

\begin{table*}[t]
\centering
\caption{Object attributes and their notations and ranges. The x/y/z values of a key point location are treated as three separate attributes. Some of the attributes are only valid under certain conditions. The attributes are used in different designed metrics. SRE stands for size relative error, GLE stands for gear location error, MRE stands for motor rotation error, SLE stands for screw location error. See a detailed explanation regarding these metrics in Section \ref{sec:regMetrics}.}
\label{table:attribute}
\resizebox{0.96\linewidth}{!}{
\begin{tabular}{ccccc}
\toprule
Notation & Attribute & Range & Validity & Involved metric \\ \midrule
$T$ & Motor type & $\{0,1,2,3,4\}$ & always & Cls. accuracy \\
$N_{s}$ & Number of cover screws & $\{3,4,5\}$ & always & Cls. accuracy \\ \midrule
$L_b$ & Bottom length & $6.2\sim 8.0$ & always & SRE \\
$L_{sb}$ & Sub-bottom length & $0.6\sim 2.0$ & always & SRE \\
$D_{lg}$ & Lower gear region diameter & $3.5\sim 4.5$ & always & SRE \\
$D_{ug}$ & Upper gear region diameter & $5.0\sim 6.5$ & if $T=0/1/2$ & SRE \\
$X_{lg}, Y_{lg}, Z_{lg}$ & Lower gear center location (xyz) & $(1.7\sim 2.3, 1.0, 10.6\sim 14.2)$ & always & $\text{GLE}_{xyz}, \text{GLE}_{xz}$ \\
$X_{ug}, Y_{ug}, Z_{ug}$ & Upper gear center location (xyz) & $(1.7\sim 2.5, 0.3\sim 0.5, 13.5\sim 17.3)$  & if $T=0/1/2$ & $\text{GLE}_{xyz}, \text{GLE}_{xz}$ \\
$R_x, R_y, R_z$ & Motor rotation (rx/ry/rz) & $(\pm 15^\circ, \pm 5^\circ, \pm 5^\circ)$ & always & MRE \\
$X_{s_1}, Y_{s_1}, Z_{s_1}$ & Cover screw 1 location (xyz) & \multirow{5}{*}{\begin{tabular}[c]{@{}c@{}}$(-4.9\sim5.0,$\\ $-3\sim-1.4,$\\ $8.6\sim20.7)$ \end{tabular}} & always & $\text{SLE}_{xyz}, \text{SLE}_{xz}$ \\
$X_{s_2}, Y_{s_2}, Z_{s_2}$ & Cover screw 2 location (xyz) &  & always & $\text{SLE}_{xyz}, \text{SLE}_{xz}$ \\
$X_{s_3}, Y_{s_3}, Z_{s_3}$ & Cover screw 3 location (xyz) &  & always & $\text{SLE}_{xyz}, \text{SLE}_{xz}$ \\
$X_{s_4}, Y_{s_4}, Z_{s_4}$ & Cover screw 4 location (xyz) &  & if $N_{s}=4/5$ & $\text{SLE}_{xyz}, \text{SLE}_{xz}$ \\
$X_{s_5}, Y_{s_5}, Z_{s_5}$ & Cover screw 5 location (xyz) &  & if $N_{s}=5$ & $\text{SLE}_{xyz}, \text{SLE}_{xz}$ \\ \bottomrule
\end{tabular}}
\end{table*}

\section{\uppercase{Synthetic dataset generation}}
\label{sec:dataset}
\subsection{Mesh model generation}
A Blender addon is created for the easy generation of synthetic motor mesh models. As an open source software, Blender \cite{bonatti2016blender} is a proven tool that performs well in modeling shapes and creating highly customizable addons. Our addon is able to generate motor mesh models with various specifications and save them in desired file formats. Each component of a generated motor can also be saved separately. 
The generated models contain following components: (i) Pole Pot; (ii) Electric Connection; (iii) Gear Container; (iv) Cover; (v) Side Screws and (vi) Cover screws. Those are the six main categories we need perform segmentation on for the first step of disassembly. Additionally, following inner components have also been generated : (vii) Magnets; (viii) Armature; (ix) Lower Gear and (x) Upper Gear, as presented in Figure \ref{fig:motorDemo2}. 
To generate motors with various specifications, we provide lots of parameter options that control the type, size, position and rotation of different parts of motor, \eg screw position, gear size, or pole pot length.
Figure \ref{fig:motorDemo1} shows ten generated demo motors with different parameters and an exploded view of a demo motor. All the individual components mentioned above are modeled separately as illustrated.

\subsection{Image and point cloud generation}
The generated mesh models are further used to create synthetic image and point cloud datasets. As shown in Figure \ref{fig:scene}, a simulated scene is built in Blender for it. Apart from the lights and cameras, to make the scene more realistic, a model of the real-world clamping system and a background panel have been added additionally. 
Three light sources with random changes in light intensity are placed in the scene.
The camera rotates randomly on top of the scene within a certain view range yet always towards the motor, as illustrated in Figure \ref{fig:camera}. To create an image dataset, apart from the scene images rendered by Blender directly, BlenderProc \cite{denninger2019blenderproc} is used to generate corresponding depth images, normal images, and segmentation ground truth images. Detection ground truth of bounding boxes of motor and screws are also provided. On the 3D synthetic data side, BlenSor \cite{gschwandtner2011blensor} is used to simulate the sensors to create point cloud data as well as to generate their segmentation ground truth. 3D bounding boxes are also given. Moreover, for the better learning of key objects, we additionally provide corresponding sub-point clouds for each scene by cropping random cuboid regions around the motors, but make sure to include them. 
The current version of our benchmark focuses more on the undismantled motors, hence inner components will not be investigated. However, any researcher interested in this part is free to use the provided addon and scripts to generate corresponding dataset according to their own needs.

\subsection{Dataset details}
We have created a synthetic motor mesh dataset of 1000 motor mesh models with different specifications. They are placed in the same simulated scene but with random camera settings, random light conditions and random mild translations and rotations. With these 1000 scenes, 1000 sets of images and 1000 point clouds are generated respectively. 80\% of the data are randomly selected as the training data, while the other 20\% are used as the test data. To ensure the correspondence between images and point clouds for each scene, the camera information has been saved and shared between BlenderProc and Blensor.

\begin{figure*}[!t]
    \centering
    \includegraphics[width=1\linewidth,trim=2 2 2 2,clip]{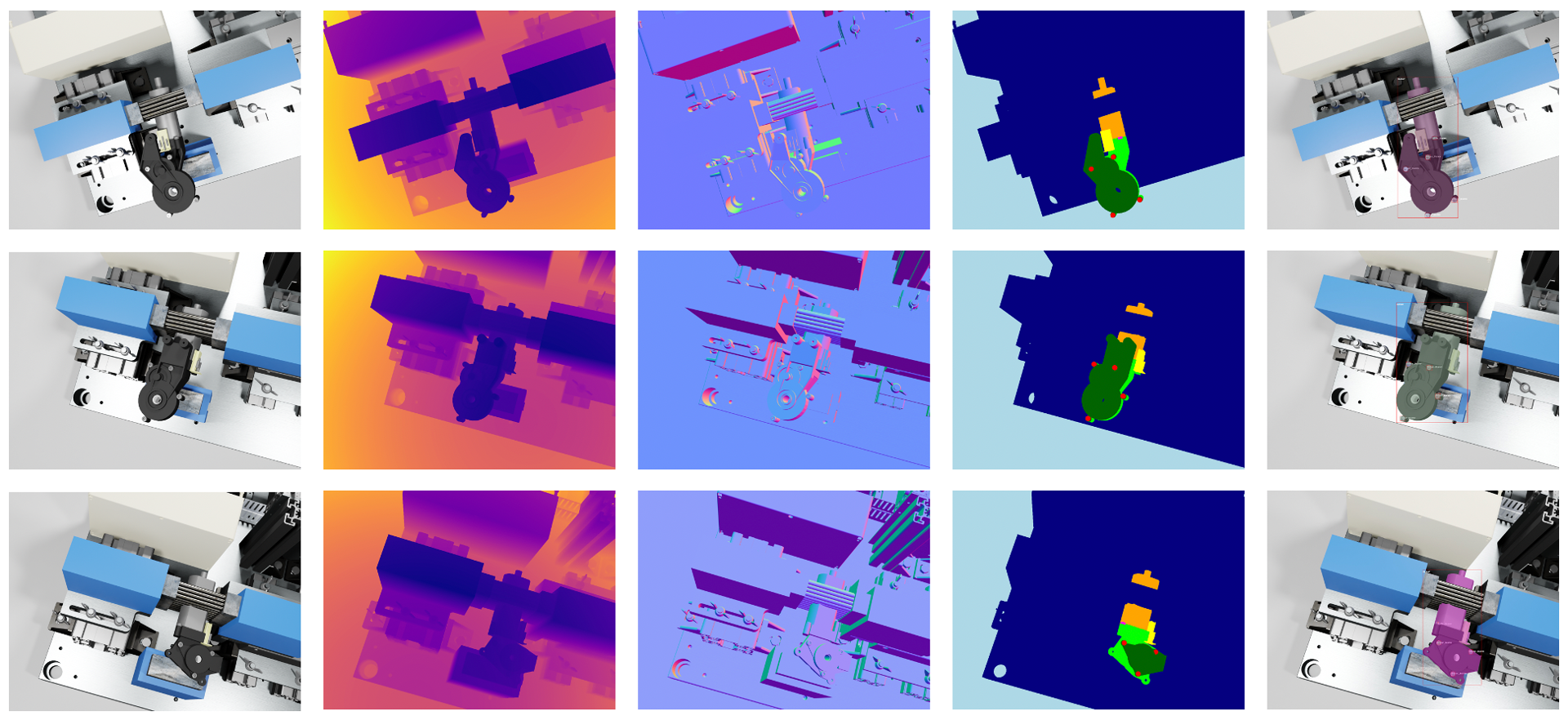}
    \caption{Demos of synthetic image data. }
    \label{fig:demoImage}
\end{figure*}

\begin{figure*}[!t]
    \centering
    \includegraphics[width=0.8\linewidth,trim=2 2 2 2,clip]{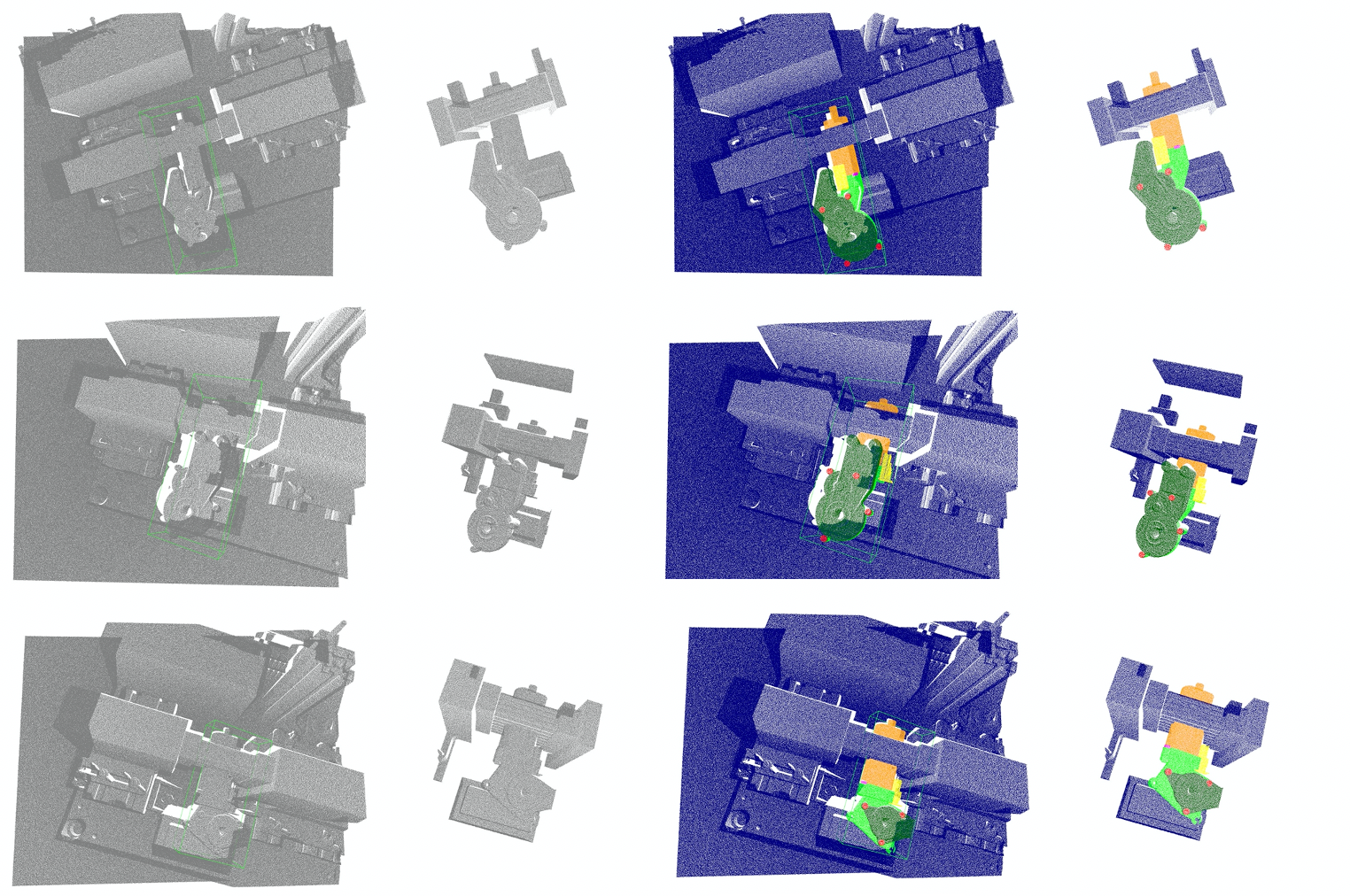}
    \caption{Demos of synthetic point cloud data. }
    \label{fig:demoPC}
\end{figure*}

Note that although we only generated 1000 motors for the dataset creation, it is possible to create a much larger dataset with more motor models following a same pipeline. Scripts are used for batch generation, hence the generation process requires no exhausting labor work at all. The only manual work is to set some hyper-parameters. The dataset generation scripts will also be released.
To give a better sense of how long it takes to generate a dataset of a certain size, we report the time it spent for creating our datasets of 1000 motors. 
To generate 1000 motor mesh models, it took around 8 hours. To generate 1000 sets of images, it took around 12 hours. To generate 1000 point clouds, it took around 21 hours. These results are based on a desktop with an Intel Core i5 CPU with 16GB RAM, and a GTX 1080 GPU. Faster processing can surely be achieved with better processors and GPUs.

Our datasets are organized as follows. 


\textbf{Mesh model dataset.} Part obj files and assembled full shape obj file are both saved, with corresponding material files. Apart from that, all attributes are saved in a csv file. We select 30 key attributes as the main learning targets for the benchmark and have pre-processed them for a more convenient usage. The 30 attributes are given in Table \ref{table:attribute}. Attribute \textit{Type} and \textit{Number of cover screws} are suitable for classification tasks, while other 28 attributes are suitable for attribute learning. Note that some attributes are invalid in some cases, \eg, the xyz coordinates of 5th cover screw when the motor only has 4 cover screws. While the attribute values are set to zero when they are invalid, an additional binary mask csv file is provided for all 28 attributes of all 1000 motors. In our benchmark, all 28 attributes are continuous variables hence are suitable for attribute regression. 


\textbf{Image dataset.} 1000 sets of images are provided. In each set, there are one rgb image, one depth image, one normal image and one segmentation ground truth image. Detection ground truth of 2D bounding boxes are provided in the COCO fashion, with a visualized detection result supplemented. A demo of generated images are given in Figure \ref{fig:demoImage}.


\textbf{Point cloud dataset.} 1000 sets of point clouds are provided. In each set, there are one scene point cloud and one cropped point cloud which focuses on the motor region. 3D bounding boxes are provided. Segmentation ground truth labels are also provided. Figure \ref{fig:demoPC} gives a demo of generated point clouds colored in their segmentation ground truth.

\section{\uppercase{Designed tasks and metrics}}
\label{sec:tasksAndMetrics}

\subsection{Common computer vision tasks}
\textbf{Detection.}
The detection task is to detect the positions of the key objects in the scene and to draw 2D or 3D bounding boxes around each object of interest in RGB images or point clouds. For 2D detection, the widely used mean average precision (mAP) was originally proposed in the VOC challenge \cite{Everingham2009ThePV}. We use an advanced version in COCO \cite{Lin2014MicrosoftCC} which further considers different IoU thresholds. The two metrics are (i) mAP with IoU threshold of $0.5$; (ii) average mAP with IoU threshold of $0.5, 0.55, 0.6, \ldots, 0.95$. For 3D detection, 
we require a 3D bounding box overlap of $70\%$ for computing the precision-recall curve and the mAP.

\textbf{Classification.}
The task of classification is to classify the key object in the scene into the prior-defined categories. There are 5 types of motors in our dataset, the main differences between them are the number of gears and the shape of covers. The classification task can be performed on both 2D dataset and 3D dataset. The well-known classification accuracy is used as the metric for the classification task. This is a relatively easy task in our setting since we only have 5 categories for classification and none of them is a tail category.

\textbf{Segmentation.}
Segmentation on 2D images is the process of assigning a label to every pixel in the image such that pixels with the same label share certain characteristics. 3D point cloud segmentation is the process of classifying point clouds into different regions, so that the points in the same isolated region have similar properties. Common segmentation metrics are category-wise IoU and overall mIoU. In our case, the overall mIoU metric is used. Moreover, since the screw categories are the key categories in real-world applications, an additional metric of screw mIoU is performed. 

\subsection{Object attribute regression}
\label{sec:regMetrics}
When generating the motor mesh models, their detailed geometric attributes are also saved, including the length of the pole pot, the diameter of the gear region, the positions of the screws, etc., which provides the possibility for regression learning. 
Evaluation metrics for assessing these learned attributes are proposed with consideration on multi-perspectives, including object size, object orientation, and object key point positions. 
It is worth noting that since the attributes are not always valid in all scenes, a binary mask is used for the metrics to get more accurate error information.
In the following definition of each metric, for a more clearly description, we use $A$ to denote the union set of involved attributes, with a corresponding binary mask $M_a$. 
Note that the following metrics are only used for evaluation, the loss used for the training is computed with a masked mean square error (MSE) between the predicted results and the ground truth.

\begin{figure}[t]
    \centering
    \includegraphics[width=0.88\linewidth]{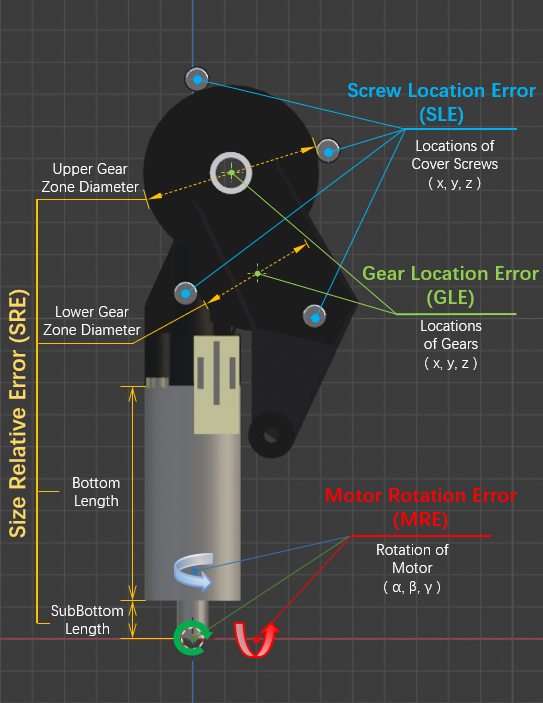}
    \caption{Object attribute regression metrics.}
    \label{fig:reg_metrics}
\end{figure}

\textbf{Size Relative Error (SRE).} This metric evaluates the predicted overall motor size using four key attributes which represent the main body of motors: the lengths of the bottom and the sub-bottom part, \ie \ pole pot, and the diameters of the gear regions. Denote $A$ as the union set of those attributes: $A\in \{L_b, L_{sb}, D_{lg}, D_{ug}\}$ with a corresponding binary mask $M_a$, for a batch of $N$ motor point clouds with ground truth $A^*$, the SRE metric is given as
\begin{equation}
\small
    \text{SRE} = \frac{1}{N}\sum_{i=1}^N \left( \frac{1}{\|M_{a_i}\|_1}\sum_{A_i} \frac{|M_{a_i} A_i - A_i^*|}{1-M_{a_i} + A_i^*} \right)
\end{equation}
where $\sum M_{a_i}$ counts the valid attribute number in $i$th motor and $1-M_{a_i}$ is served as a safety term to ensure the denominator is greater than $0$ When $A_i^*=0$. For a certain attribute of a certain motor that is invalid, its ground truth value and mask value are both $0$, \ie \ $A_i^* = M_{a_i} = 0$, its size relative error is computed as zero and this attribute will not be counted in the denominator.

Note that we compute a mean error over the instance level for SRE. We believe this can represent the metric information better. The following three metrics are performed over the batch level.

\textbf{Gear Location Error (GLE).} This metric evaluates the distance error between the predicted location of gear region center and its ground truth. Involved attributes are the center point coordinate values. Denote $A_l$ and $A_u$ as the union sets of those attributes: $A_l\in \{X_{lg}, Y_{lg}, Z_{lg}\}$ and $A_u\in \{X_{ug}, Y_{ug}, Z_{ug}\}$ with masks $M_{al}$ and $M_{au}$, for a batch of $N$ motor point clouds, the $\text{GLE}_{xyz}$ metric is given as

\begin{equation}
\resizebox{1\linewidth}{!}{$
    \text{GLE}_{xyz} = \frac{\sum_{i=1}^N \left(\sqrt{\sum_{A_{l_i}} \left(M_{al_i} A_{l_i} - A_{l_i}^*\right)^2} + \sqrt{\sum_{A_{u_i}}\left(M_{au_i} A_{u_i} - A_{u_i}^*\right)^2} \ \right)}{\sum_{i=1}^N \left( \|M_{al_i}\|_1/3 +  \|M_{au_i}\|_1/3\right)}
    $}
\end{equation}
where $M_{au_i}/3$ means for the 3D coordinate of each center point, the mask should only be counted for one time.
In real-world applications of disassembly, the small error along the motor normal direction (in our case, the Y axis) is sometimes irrelevant. We further provide another metric that only considers the distance error on the XZ plane with projected points. By redefining $A_l\in \{X_{lg}, Z_{lg}\}$ and $A_u\in \{X_{ug}, Z_{ug}\}$, a similar metric of $\text{GLE}_{xz}$ is given as
\begin{equation}
\resizebox{1\linewidth}{!}{$
    \text{GLE}_{xz} = \frac{\sum_{i=1}^N \left(\sqrt{\sum_{A_{l_i}} \left(M_{al_i} A_{l_i} - A_{l_i}^*\right)^2} + \sqrt{\sum_{A_{u_i}} \left(M_{au_i} A_{u_i} - A_{u_i}^*\right)^2} \ \right)}{\sum_{i=1}^N \left( \|M_{al_i}\|_1/2 + \|M_{au_i}\|_1/2\right)}
    $}
\end{equation}

\textbf{Motor Rotation Error (MRE).} This metric evaluates the absolute motor rotation error. Involved attributes are the motor rotations along three axes. Denote $A$ as the union set of those involved attributes: $A\in \{R_x, R_y, R_z\}$. Since the rotation attributes are always valid, the mask $M_a$ is unnecessary for computation. The metric MRE is defined as
\begin{equation}
\small
    \text{MRE} = \frac{1}{3N}\sum_{i=1}^N \sum_{A_i} |A_i - A_i^*| 
\end{equation}

\textbf{Screw Location Error (SLE).} This metric evaluates the distance error between the predicted cover screw positions and their ground truth. Involved attributes are the screw position coordinate values. Denote $A_j$ as the union set of those attributes: $A_j\in \{X_{s_j}, Y_{s_j}, Z_{s_j}\}$ with masks $M_{aj}$ where $j=1,2,3,4,5$. For a batch of $N$ motor point clouds, the $\text{SLE}_{xyz}$ metric is given as
\begin{equation}
\small
    \text{SLE}_{xyz} = \frac{\sum_{i=1}^N \sum_{j=1}^5 \sqrt{\sum_{A_{j_i}} \left(M_{aj_i} A_{j_i} - A_{j_i}^*\right)^2} }{\sum_{i=1}^N \sum_{j=1}^5 \|M_{aj_i}\|_1/3}
\end{equation}
where $M_{aj_i}/3$ means for the 3D coordinate of each screw, the mask should only be counted for one time. Same as GLE, a metric $\text{SLE}_{xz}$ only considers the XZ plane is defined with $A_j\in \{X_{s_j}, Z_{s_j}\}$:
\begin{equation}
\small
    \text{SLE}_{xz} = \frac{\sum_{i=1}^N \sum_{j=1}^5 \sqrt{\sum_{A_{j_i}} \left(M_{aj_i} A_{j_i} - A_{j_i}^*\right)^2} }{\sum_{i=1}^N \sum_{j=1}^5 \|M_{aj_i}\|_1/2}
\end{equation}

\subsection{Multi-task learning}
Multi-task learning \cite{Ruder2017AnOO}\cite{Zhang2017ASO} means solving multiple learning tasks simultaneously, while exploiting the commonalities and differences between tasks. This can improve the learning efficiency and prediction accuracy of task-specific models compared to training the models individually. While acceptable performance can be obtained by focusing on a single task, the information that might help in getting better performance is possibly ignored. Specifically, the information comes from training signals on related tasks. By sharing representations between related tasks, the generalization ability of the model can be improved on the original task. 

In our case, the above tasks can be performed simultaneously with a same backbone network. For example, classification and segmentation can be performed at the same time. Or as mentioned in the last subsection, the classification, detection, and segmentation results may be used as additional input for the regression task. Multi-modal learning is also possible by using both 2D and 3D data.

\begin{table*}[t]
\caption{3D classification and segmentation baseline results}
\label{tab:cls_seg}
\centering
\resizebox{1.0\linewidth}{!}{
\begin{tabular}{ccclcccccccccl}
\toprule
\multirow{2}{*}{Model} & \multirow{2}{*}{Loss} & Classification &  & \multicolumn{9}{c}{Segmentation (mIoU)} &  \\ \cmidrule{3-3} \cmidrule{5-13}
 &  & (Accuracy) &  & Overall & Background & Cover & Gear Container & Charger & \multicolumn{1}{l}{Bottom} & Side Screw & Cover Screw & Screw &  \\ \midrule
DGCNN (cls) & $L_c$ & 100 &  & - & - & - & - & - & - & - & - & - &  \\
DGCNN (seg) & $L_s$ & - &  & 93.24 & 99.89 & 98.63 & 95.80 & 97.32 & 98.29 & 79.72 & 83.10 & 82.73 &  \\
DGCNN (cls+seg) & $L_c+L_s$ & 100 &  & 92.47 & 99.87 & \multicolumn{1}{l}{98.43} & 95.41 & 96.97 & 98.15 & 77.05 & 81.38 & 80.83 &  \\
DGCNN (cls+seg) & $\sqrt{L_{c}L_{s}}$ & 100 &  & 92.41 & 99.87 & 98.40 & 95.35 & 96.87 & 98.18 & 76.94 & 81.20 & 80.66 &  \\ \bottomrule
\end{tabular}}
\end{table*}

\begin{table}[t]
\caption{2D Detection baseline results.}
\label{table:detection}
\centering
\scalebox{0.85}{
\begin{tabular}{ccc}
\toprule
Model & mAP (IoU 0.5-0.95) & mAP (IoU 0.5) \\ \midrule
YOLOv5n & 75.3 & 93.6 \\
YOLOv5s & 77.8 & 94.2 \\
YOLOv5m & 82.7 & 95.9 \\
YOLOv5l & 86.7 & 96.4 \\
YOLOv5x & 89.1 & 96.8 \\\bottomrule
\end{tabular}}
\end{table}

\section{\uppercase{Baseline results}}
\label{sec:baseline}

\subsection{2D detection}
Since there is no such network architecture for 3D point cloud detection as widely recognized as YOLO series for 2D image detection, for the detection task, we give baseline results on the 2D image dataset. YOLO models of different sizes are used. All the experiments are performed with same parameter settings. The input image resolution is 640, the batch size is 16. The optimizer SGD is used with an epoch number of 200. The learning rate starts at 0.01 and decays to 0.001 with a linear decay.
The mAP performances of all models are given in Table \ref{table:detection}. It shows that the YOLO framework achieves remarkable performance on detection tasks. The mAP performance also improves when a larger network model is used.

\subsection{3D classification and segmentation}
In the past five years, a variety of network models have been proposed for point cloud data.
Multi-view based \cite{lawin2017deep}\cite{boulch2017unstructured} and volumetric-based methods \cite{maturana2015voxnet}\cite{jiang2018pointsift} are mostly used in the early years. Since the pioneer work of PointNet \cite{Qi2017PointNetDL}, point-based methods which include point-wise MLP \cite{qi2017pointnet++}, point convolution-based methods \cite{Wu2019PointConvDC}\cite{thomas2019kpconv} and graph-based methods \cite{wang2019dynamic}\cite{Chen2021GAPNetGA}\cite{Liang20203DIE}, gradually became the main choice. Recent work even adapt the idea of transformer \cite{vaswani2017attention} for point cloud learning, \eg, PCT \cite{Guo2021PCTPC} and PT \cite{Zhao2020PointT}\cite{Engel2021PointT}. Among all those methods, PointNet++ \cite{qi2017pointnet++} and DGCNN \cite{wang2019dynamic} are recognized as two key work in the domain. 
In this paper, we use DGCNN as the network backbone to produce baseline results.

\begin{figure}[t]
    \centering
    \includegraphics[width=1\linewidth,trim=2 2 2 2,clip]{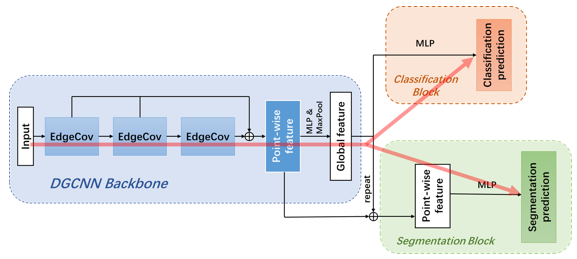}
    \caption{3D classification and segmentation framework for baseline results. \vspace{-0.0cm}}
    \label{fig:multiTask}
\end{figure}

We use an architecture of three successive EdgeConv blocks (key block in DGCNN) to learn point-wise features, a linear layer and a max pooling layer is applied subsequently to compute a global feature. 
For the classification task, the global feature is processed with several followed dense layers to get a classification prediction; while for the segmentation task, the global feature is repeated and concatenated with point-wise features to process through several other dense layers to predict point-wise labels.
An illustrative figure is given in Figure \ref{fig:multiTask}. Note that although two tail blocks are different, the encoder block is identical. It is possible to use a shared encoder for both classification and segmentation tasks. In this case, by adding the losses together and performing gradient back propagation at the same time, we can train on the classification task and the segmentation task simultaneously. This is a simple way of performing multi-task learning. The red line indicates how the information flows during one training step. The encoder block is co-trained with both tasks, while the respective task tail blocks are trained in parallel.

The settings of these three experiments are identical. We use a sub-point cloud size of 2048 points for batch training. The batch size is 16. The optimizer AdamW is used with an epoch number of 100. The learning rate starts with $1\times 10^{-3}$ and decays to $1\times 10^{-5}$ with a cosine annealing schedule. In the EdgeConv blocks, we use $K=32$ when selecting neighbor points.
Table \ref{tab:cls_seg} indicates that the DGCNN global feature allows for a complete identification of the motor types since it is a relatively easy task. However, the segmentation results from the multi-task training model are not as good as that from individual training. 
Using loss weights to balance the gradient information from two tails blocks may improve the results. We would like to leave this problem for other researchers that are interested in this topic.

\begin{table*}[h]
\caption{3D regression baseline results. $T$ stands for motor type, $N_s$ stands for the number of cover screws. SRE stands for size relative error, GLE stands for gear location error, MRE stands for motor rotation error, SLE stands for screw location error. Results on all four kinds of methods are presented. Results with or without using the additional segmentation task in the meta-block are both presented.}
\label{tab:regression}
\centering
\resizebox{1\linewidth}{!}{
\begin{tabular}{ccclclclclclclclclclcc}
\toprule
\multirow{2}{*}{Method} & \multicolumn{1}{l}{\multirow{2}{*}{with seg}} & \multicolumn{3}{c}{Classification $\uparrow$} &  & \multicolumn{3}{c}{Segmentation $\uparrow$} &  & \multicolumn{1}{l}{} &  & \multicolumn{1}{l}{} & \multicolumn{3}{c}{Regression $\downarrow$} & \multicolumn{1}{l}{} &  & \multicolumn{1}{l}{} &  & \multicolumn{1}{l}{} &  \\ \cmidrule{3-5} \cmidrule{7-9} \cmidrule{11-21}
 & \multicolumn{1}{l}{} & \multicolumn{1}{l}{$T$ accuracy} &  & \multicolumn{1}{l}{$N_s$ accuracy} &  & mIoU &  & \multicolumn{1}{l}{Screw mIoU} &  & SRE &  & $\text{GLE}_{xyz}$ &  & $\text{GLE}_{xz}$ &  & MRE &  & $\text{SLE}_{xyz}$ &  & $\text{SLE}_{xz}$ &  \\ \midrule
\multirow{2}{*}{Separate} & no & 100 &  & 80.86 &  & - &  & - &  & 6.20\% &  & 0.3745 &  & 0.3741 &  & 0.9398 &  & 0.6171 &  & 0.6138 &  \\
 & yes & 100 &  & 82.52 &  & 90.27 &  & 74.84 &  & 6.20\% &  & 0.3745 &  & 0.3741 &  & 0.9398 &  & 0.6171 &  & 0.6138 &  \\
\multicolumn{1}{l}{\multirow{2}{*}{Pre-train}} & no & 100 &  & 80.86 &  & - &  & - &  & 5.56\% &  & 0.3423 &  & 0.3418 &  & 0.8237 &  & 0.5615 &  & 0.5585 &  \\
\multicolumn{1}{l}{} & yes & 100 &  & 82.52 &  & 90.27 &  & 74.84 &  & \textbf{5.41\%} &  & 0.3500 &  & 0.3499 &  & \textbf{0.8039} &  & 0.5556 &  & 0.5536 &  \\
\multirow{2}{*}{Parallel} & no & 100 &  & 75.98 &  & - &  & - &  & 5.66\% &  & \textbf{0.3199} &  & \textbf{0.3196} &  & 0.8233 &  & \textbf{0.5500} &  & \textbf{0.5477} &  \\
 & yes & 100 &  & 76.40 &  & 88.65 &  & 71.36 &  & 5.81\% &  & 0.3759 &  & 0.3757 &  & 0.8518 &  & 0.5869 &  & 0.5829 &  \\
\multirow{2}{*}{Iterative} & no & 100 &  & 81.52 &  & \textbf{-} &  & \textbf{-} &  & 6.03\% &  & 0.4043 &  & 0.4040 &  & 0.9386 &  & 0.6100 &  & 0.6053 &  \\
 & yes & 100 &  & \textbf{82.82} &  & \textbf{90.49} &  & \textbf{75.63} &  & 5.84\% &  & 0.3721 &  & 0.3719 &  & 0.9444 &  & 0.5759 &  & 0.5740 &  \\ \bottomrule
\end{tabular}}
\end{table*}

\begin{figure*}[t]
\begin{subfigure}{.5\textwidth}
    \centering
    \includegraphics[width=.85\linewidth]{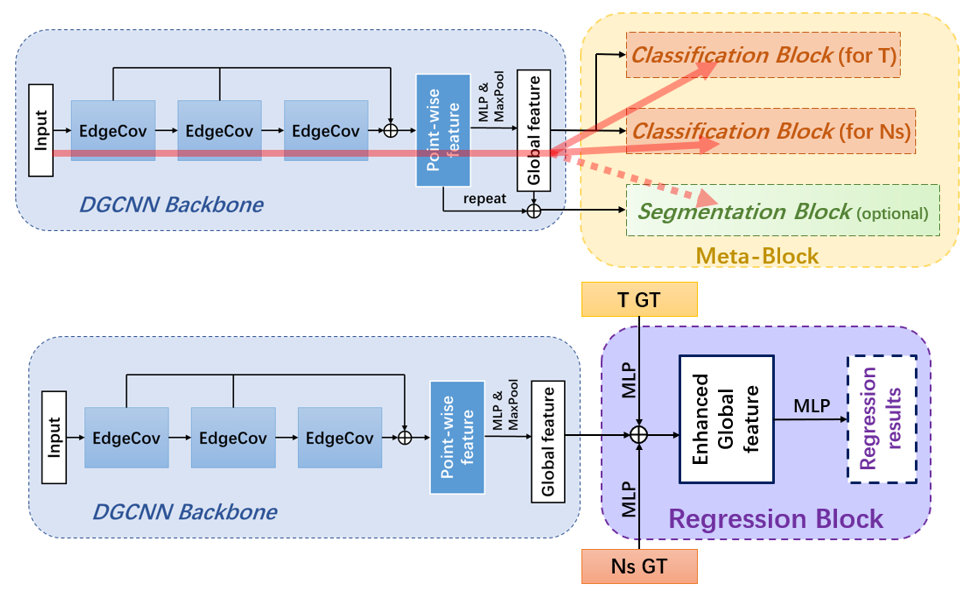}
    \caption{Totally separate training.\vspace{0.2cm}}
    \label{separate}
\end{subfigure}
\begin{subfigure}{.5\textwidth}
    \centering
    \includegraphics[width=.95\linewidth]{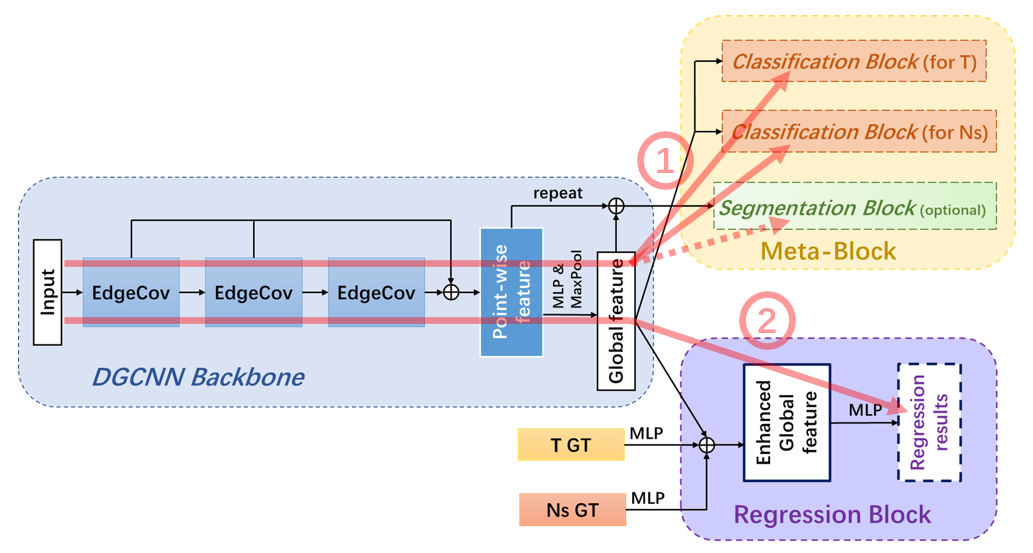}
    \caption{Use meta-block for pre-training.\vspace{0.2cm}}
    \label{pretrain}
\end{subfigure}
\begin{subfigure}{.5\textwidth}
    \centering
    \includegraphics[width=.95\linewidth]{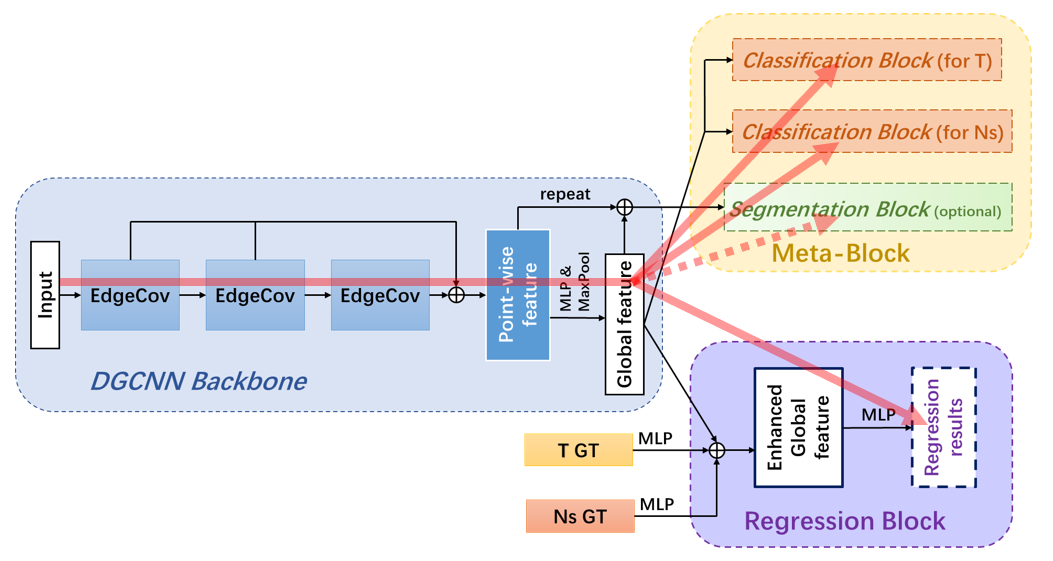}
    \caption{Encoder-shared yet tail blocks trained in parallel.\vspace{0.1cm}}
    \label{parallel}
\end{subfigure}
\begin{subfigure}{.5\textwidth}
    \centering
    \includegraphics[width=.95\linewidth]{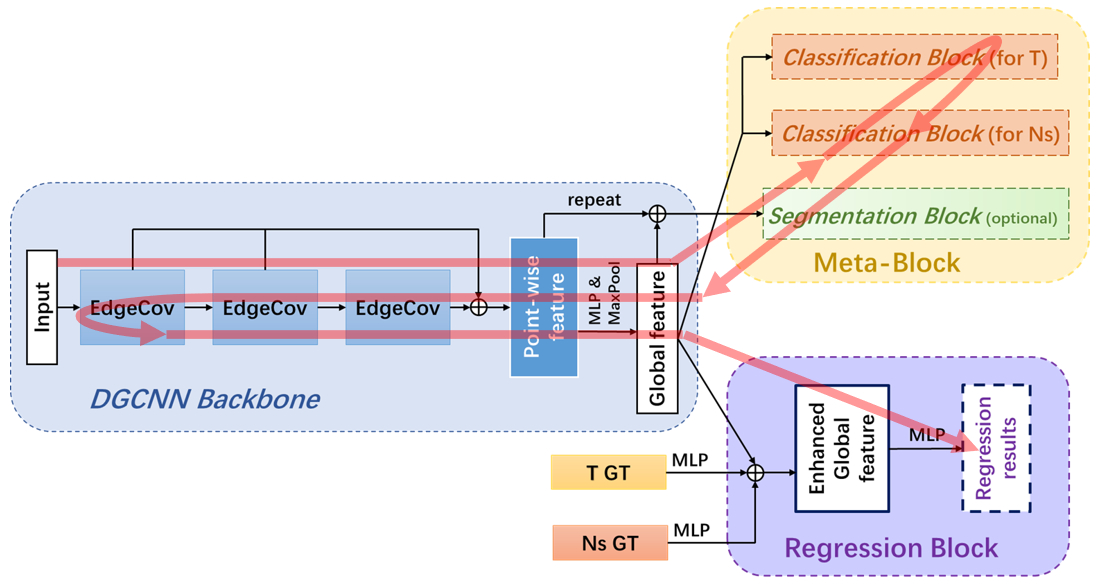}
    \caption{Encoder-shared and tail blocks trained iteratively.\vspace{0.2cm}}
    \label{iterative}
\end{subfigure}
\caption{Four different training methods for baseline results on the 3D regression framework. The red lines indicate how the information flows inside these frameworks.}
\label{fig:regression}
\end{figure*}

\subsection{3D object attribute regression}
\label{sec:regression}


Same as in the previous subsection, DGCNN is used as the backbone for encoding point-wise features and a global feature. Moreover, as illustrated in Figure \ref{fig:regression}, we consider the classification-segmentation parallel training introduced in the previous subsection as a meta-block. Note that since the one-hot attributes $T$ and $N_s$ indicate the validity of some other attributes, apart from the block of motor type classification, another block for classifying the number of cover screws has also been included in the meta-block. The regression tasks compose a second tail block. In our case, we use a simple method of concatenating the global feature with one-hot features that have been encoded with MLPs, and then directly perform MLP on the enhanced global feature to get a 28-dimensional vector, which indicates the regression results of 28 attributes.

With the proposed architecture, there are several possible ways to train those blocks. (i) \textbf{Totally separate training:} the encoder and the meta-block are trained as one network first, and then another same-structure encoder and the regression block are trained as another totally separate network. No information is shared between two trainings. (ii) \textbf{Use meta-block for pre-training:} this method is similar to the last one, but the encoder weights in the second step will be initialized with the weights from the first step. The meta-block is used for pre-training. (iii) \textbf{Encoder-shared yet tail blocks trained in parallel:} the encoder is shared between two tail blocks and three blocks compose one joint network. All tasks are trained in parallel. (iv) \textbf{Encoder-shared and tail blocks trained iteratively:} in each training step, the encoder is firstly connected with the meta-block. We compute the loss, perform gradient back propagation and update model weights with these two blocks. Then the same encoder connects with the regression block in a switch manner, the input is reprocessed with the weight-updated encoder to get new encoded representations which are used for computing the regression loss. We then again perform gradient back propagation and weight update in these two blocks. This action performs iteratively. The red line in Figure \ref{fig:regression} indicates how the information flows during one training step. 

The experiments of four methods used same settings. We set a sub-point cloud size of 2048 points for batch training. The batch size is 16. The optimizer AdamW is used with an epoch number of 100. The learning rate starts with $1\times 10^{-3}$ and decays to $1\times 10^{-5}$ with a cosine annealing schedule. In the EdgeConv blocks, we use $K=32$ when selecting neighbor points.
As shown in Table \ref{tab:regression}, in most cases, performing attribute regression with semantic segmentation decreases the regression error. Regression can also help in determining the number of cover screws. 
Among them, the separate-based method is the most simple one and it achieves the worst regression results. The pretrain-based method significantly improves the regression results. 
Overall, the parallel-based method achieves better performance in most regression metrics. However, since losses are merged together during the training, the classification and segmentation performances drop slightly with the parallel-based method.
The iterative training-based approach achieves the best results for segmentation, but does not achieve the best results in regression.
There are surely some other better ways of designing the architecture, \eg, post-processing on the segmentation results for better $N_s$ classification results, or using 2D detection results as the supplementary input. We would like to leave this as an open question for other interested researchers.

\section{\uppercase{Conclusion}}
\label{sec:conclusion}
In this paper, a benchmark is proposed using synthetic 2D and 3D dataset, in which motors are the key objects. Motor attributes are saved during the dataset generation and are suitable for the less explored attribute regression task. Apart from the common computer vision tasks including classification, detection, and segmentation, several metrics have been designed especially for the regression task. Baseline results on several tasks are also provided. We hope this work could contribute to the attribute regression or multi-task learning domain in the computer vision community and inspire the development of other novel algorithms in the future.

\section*{Acknowledgements}
The project AgiProbot is funded by the Carl Zeiss Foundation.

\bibliographystyle{apalike}
{\small
\bibliography{example}}

\end{document}